\documentclass[12pt]{article}
\usepackage{graphicx}
\usepackage{amsmath}
\usepackage{lipsum}
\usepackage{hyperref}
\usepackage{enumitem}
\usepackage{float}
\usepackage{enumitem}

\title{\textbf{Multi Part Deployment of Neural Network}}
\author{
 Paritosh Ranjan \\
  IBM  \\
  \texttt{paranjan@in.ibm.com} \\
  \and
 Surajit Majumder \\
  IBM  \\
  \texttt{surajit.majumder@ibm.com} \\
  \and
 Prodip Roy \\
  IBM  \\
  \texttt{prodipro@in.ibm.com} \\
}
\date{\today}

\begin{document}

\maketitle

\begin{abstract}
The increasing scale of modern neural networks, exemplified by architectures from IBM (530 billion neurons) and Google (500 billion parameters), presents significant challenges in terms of computational cost and infrastructure requirements. As deep neural networks continue to grow, traditional training paradigms relying on monolithic GPU clusters become increasingly unsustainable. This paper proposes a distributed system architecture that partitions a neural network across multiple servers, each responsible for a subset of neurons. Neurons are classified as local or remote, with inter-server connections managed via a metadata-driven lookup mechanism. A Multi-Part Neural Network Execution Engine facilitates seamless execution and training across distributed partitions by dynamically resolving and invoking remote neurons using stored metadata. All servers share a unified model through a network file system (NFS), ensuring consistency during parallel updates. A Neuron Distributor module enables flexible partitioning strategies based on neuron count, percentage, identifiers, or network layers. This architecture enables cost-effective, scalable deployment of deep learning models on cloud infrastructure, reducing dependency on high-performance centralized compute resources.

\end{abstract}

\section{Introduction}

The rapid expansion in the scale of neural networks has been a defining trend in recent deep learning research, fueled by innovations in model architectures and the proliferation of large-scale datasets. Prominent milestones include IBM’s brain-inspired architecture—featuring approximately 530 billion neurons and 100 trillion synapses trained on supercomputing resources—and Google’s neural network models with up to 500 billion parameters. These examples underscore an ongoing shift toward ever-larger and more complex deep neural networks (DNNs).

Despite their impressive capabilities, the training and deployment of such large-scale models demand substantial computational and financial resources. High-performance computing clusters equipped with GPUs are often required, resulting in significant infrastructure costs that can hinder broader adoption and experimentation. This challenge is particularly pronounced in cloud environments, where scalability and cost efficiency are critical considerations.

In this work, we address the pressing challenge of enabling scalable, cost-effective training and deployment of large neural networks in cloud-based settings. We introduce a distributed neural network architecture that partitions computational workloads across multiple servers, thereby reducing infrastructure expenses while maintaining computational accuracy. Our proposed approach facilitates the scalable deployment of deep learning models without reliance on monolithic compute nodes, advancing the accessibility and efficiency of state-of-the-art AI systems.

\section{Brief Description of the Invention}

The proposed invention introduces a distributed framework for both training and running deep neural networks by segmenting the network across several servers. Each server is responsible for a portion of the network’s neurons—termed local neurons—and retains metadata about remote neurons, which reside on other servers but are interconnected with the local neurons. This metadata encompasses information such as neuron identifiers, associated weights, and the details of the remote servers.

At the core of this system is the \textbf{Multi-Part Neural Network Execution Engine}, which manages both training and inference processes by determining whether a neuron is local or remote. For remote neurons, the engine accesses the relevant metadata and initiates a remote procedure call to the server hosting that neuron, enabling the required computations. All servers collaborate using a shared model housed on a \textbf{network file system} (NFS), which ensures consistency and synchronization of model updates.

Neuron distribution across servers is handled by a \textbf{Neuron Distributor} component, which offers various allocation strategies, including fixed numbers, percentages, specific network layers, or unique identifiers. This flexible and scalable architecture supports efficient execution of large-scale neural networks in cloud environments, substantially lowering the costs and complexity compared to traditional centralized GPU cluster solutions.

\section{Reduction to Practice}

The invention has been practically realized through the development of a distributed neural network system where neurons are divided among several servers. Each server operates a Multi-Part Neural Network Execution Engine, which utilizes metadata to distinguish and process both local and remote neurons, facilitating synchronized training and inference throughout the distributed setup. All servers update a shared model stored on a network file system, guaranteeing both consistency and scalability.

This invention is made up of following components:

\subsection{Local Neurons}

Local neurons refer to those Neurons which are present within the current server/current partition where the computation is being done (e.g., Server 1)

\subsection{Remote Neurons}

Remote neurons refer to those Neurons that are present on other servers/other partitions and connected with Neurons on current server/current partition. 

To monitor these connections, each server is equipped with a lookup mechanism at its entry gate. This lookup mechanism is responsible for determining whether neurons are local or remote by checking the stored metadata.

\subsection{Neuron Metadata}

On every server there is a local storage which stores Neuron metadata ONLY about the connections between the local neurons and the connected remote neurons. The Neuron Metadata would contain the server information server IP, server credentials, Neuron identifier, weight of the connection etc, for each remote Neuron which is connected with a local Neuron on the current server.

\subsection{Multi Part Neural Network Execution Engine}

The Multi Part Neural Network Execution Engine would execute both Local Neurons and Remote Neurons performing the training and all Neural Network computations. 
For every Neuron it will check in the Neuron Metadata, if the Neuron is a local Neuron or a remote Neuron.
If the Neuron is a remote Neuron, then this engine would extract the Neuron Metadata of the remote Neuron and call the Remote Neuron from the respective current Neuron. The call will be received by the Multi Part Neural Network Execution Engine on the remote server containing the remote Neuron and would be taken forward for computation in the same fashion.

\subsection{Shared Storage}

The systems would share storage via Network File System.

\subsection{Single Model}

All servers would write to the same model on the shared storage. So, there would be only one model all the time in this system.

\subsection{Neuron Distributor}
There would be a Neuron Distributor component which would distribute neurons to different partition servers. 

There would be multiple distribution mechanisms:
\begin{enumerate}
    \item Defined percentage of neurons in each partition with neurons selected randomly or from beginning or end of a sorted list of neurons
    \item Defined count of neurons in each partition with neurons selected randomly or from beginning or end of a sorted list of neurons
    \item Unique identifiers of neurons in each partition
    \item Specific Layers of neural network in each partition
\end{enumerate}

\begin{figure}[h!]
    \centering
    \includegraphics[width=0.8\textwidth]{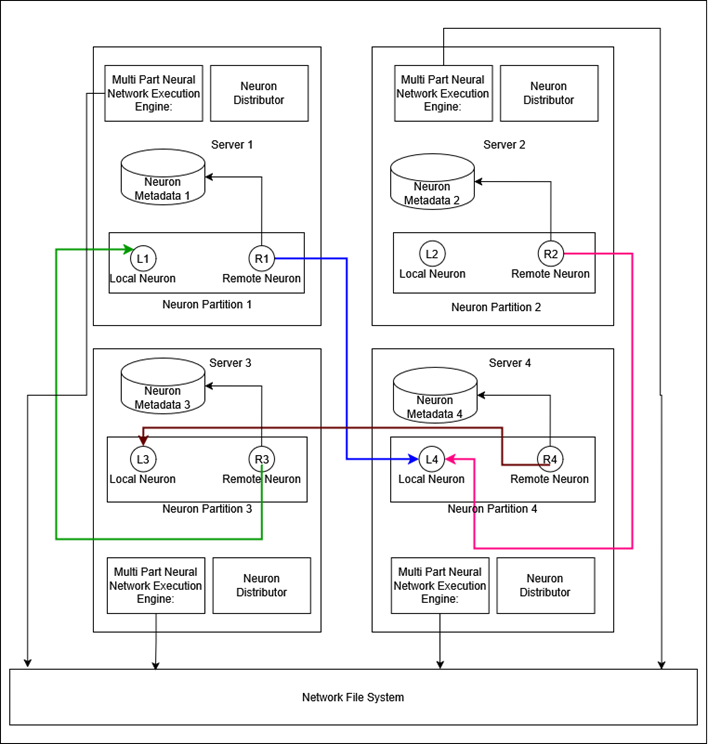}
    \caption{Component-Diagram}
    \label{fig:Component-Diagram}
\end{figure}

\section{Conclusion}
The increasing size and complexity of neural networks have made traditional, centralized training methods increasingly impractical due to high computational costs and infrastructure demands. This work introduces a distributed architecture that effectively partitions a neural network across multiple servers, allowing each server to process a designated subset of neurons. Through the use of neuron metadata and a coordinated execution engine, the system ensures seamless interaction between local and remote neurons during both training and inference. By leveraging shared storage and distributed computation, the proposed solution offers a cost-efficient, scalable approach suitable for cloud environments. Moving forward, enhancements in communication efficiency and system robustness could further improve performance and adaptability.

\section{Acknowledgment}

We would like to express our sincere gratitude to all individuals and organizations who have contributed to the success of this research. We acknowledge the invaluable support from the IBM team, whose resources and expertise have greatly enhanced this project.
Special thanks to Prodip Roy (Program Manager IBM) for their insightful feedback, guidance, and encouragement throughout the development of this work.
\section{References}
\renewcommand\refname{}

\end{document}